\documentclass[10pt,twocolumn,letterpaper]{article}

\usepackage{ijcb}
\usepackage{times}
\usepackage{epsfig}
\usepackage{graphicx}
\usepackage{amsmath}
\usepackage{amssymb}
\usepackage{multirow}
\usepackage{makecell}
\usepackage{setspace}

\usepackage[norule,symbol,perpage]{footmisc}

\ijcbfinalcopy % *** Uncomment this line for the final submission

\ifijcbfinal\fi

\makeatletter
\def\ps@IEEEtitlepagestyle{
\def\@oddfoot{\mycopyrightnotice}
\def\@evenfoot{}
}

\def\thanks#1{\protected@xdef\@thanks{\@thanks
        \protect\footnotetext{#1}}}

\def\mycopyrightnotice{
{\hfill \footnotesize 978-1-6654-3780-6/21/\$31.00 \copyright 2021 IEEE\hfill}
}
\makeatother

\begin{document}

%%%%%%%%% TITLE
\title{Face Sketch Synthesis via Semantic-Driven Generative Adversarial Network}

\author{Xingqun Qi$^{1, 2, \ast}$\thanks { $\ast$ These authors contributed equally to this work. This work was done when Xingqun Qi was an intern in the CRIPAC, NLPR, CASIA.}
, Muyi Sun$^{2, \ast}$
, Weining Wang$^{2, \dagger}$\thanks { $\dagger$ Corresponding authors.}, Xiaoxiao Dong$^{2, 3}$, Qi Li$^{2}$, Caifeng Shan$^{3,4, \dagger}$\\
$^{1}$School of Automation, Beijing University of Posts and Telecommunications, Beijing, China\\
$^{2}$Center for Research on Intelligent Perception and Computing, NLPR, CASIA, Beijing, China\\
$^{3}$Artificial Intelligence Research, CAS, Jiaozhou, Qingdao, China\\
$^{4}$Shandong University of Science and Technology, Qingdao, China\\
{\tt\small xingqunqi@bupt.edu.cn, \{muyi.sun, xiaoxiao.dong\}@cripac.ia.ac.cn  } \\
{\tt\small \{qli, weining.wang\}@nlpr.ia.ac.cn, caifeng.shan@gmail.com}

%\and
}

\maketitle

\thispagestyle{empty}

%%%%%%%%% ABSTRACT
\begin{abstract}
Face sketch synthesis has made significant progress with the development of deep neural networks in these years. The delicate depiction of sketch portraits facilitates a wide range of applications like digital entertainment and law enforcement. However, accurate and realistic face sketch generation is still a challenging task due to the illumination variations and complex backgrounds in the real scenes. To tackle these challenges, we propose a novel Semantic-Driven Generative Adversarial Network (SDGAN) which embeds global structure-level style injection and local class-level knowledge re-weighting. Specifically, we conduct facial saliency detection on the input face photos to provide overall facial texture structure, which could be used as a global type of prior information. In addition, we exploit face parsing layouts as the semantic-level spatial prior to enforce globally structural style injection in the generator of SDGAN. Furthermore, to enhance the realistic effect of the details, we propose a novel Adaptive Re-weighting Loss (ARLoss) which dedicates to balance the contributions of different semantic classes. Experimentally, our extensive experiments on CUFS and CUFSF datasets show that our proposed algorithm achieves state-of-the-art performance.
\end{abstract}

\let\thefootnote\relax\footnotetext{\mycopyrightnotice}

%%%%%%%%% BODY TEXT
\section{Introduction}

Face sketch synthesis is a critical and challenging task in computer vision which refers to generating sketches from face photos. Face sketch synthesis attracts plentiful attention in recent years and plays a significant role in a wide range of applications like digital entertainment and law enforcement. In digital entertainment, face sketch is one of the most fundamental and popular portrait styles. Nevertheless, it requires vast time and efforts to create realistic face sketches by professional artists. Meanwhile, in law enforcement and criminal justice cases, clearly identifiable photos of criminals are often tough to obtain. In this case, face sketches with manifest features could provide a feasible way to promote the efficiency of the police \cite{wang2014comprehensive}. Therefore, it is especially essential to automatically synthesize facial sketches with realistic effects and preserve identity features. So far, numerous approaches have been proposed to facilitate face sketch synthesis. Generally speaking, these methods can be roughly divided into three categories: exemplar-based approaches \cite{tang2003face}, linear-regression based approaches \cite{zhu2016simple} and model-based approaches \cite{long2015fully}.

\begin{figure}[t]
\begin{center}
\includegraphics[width=0.85\linewidth]{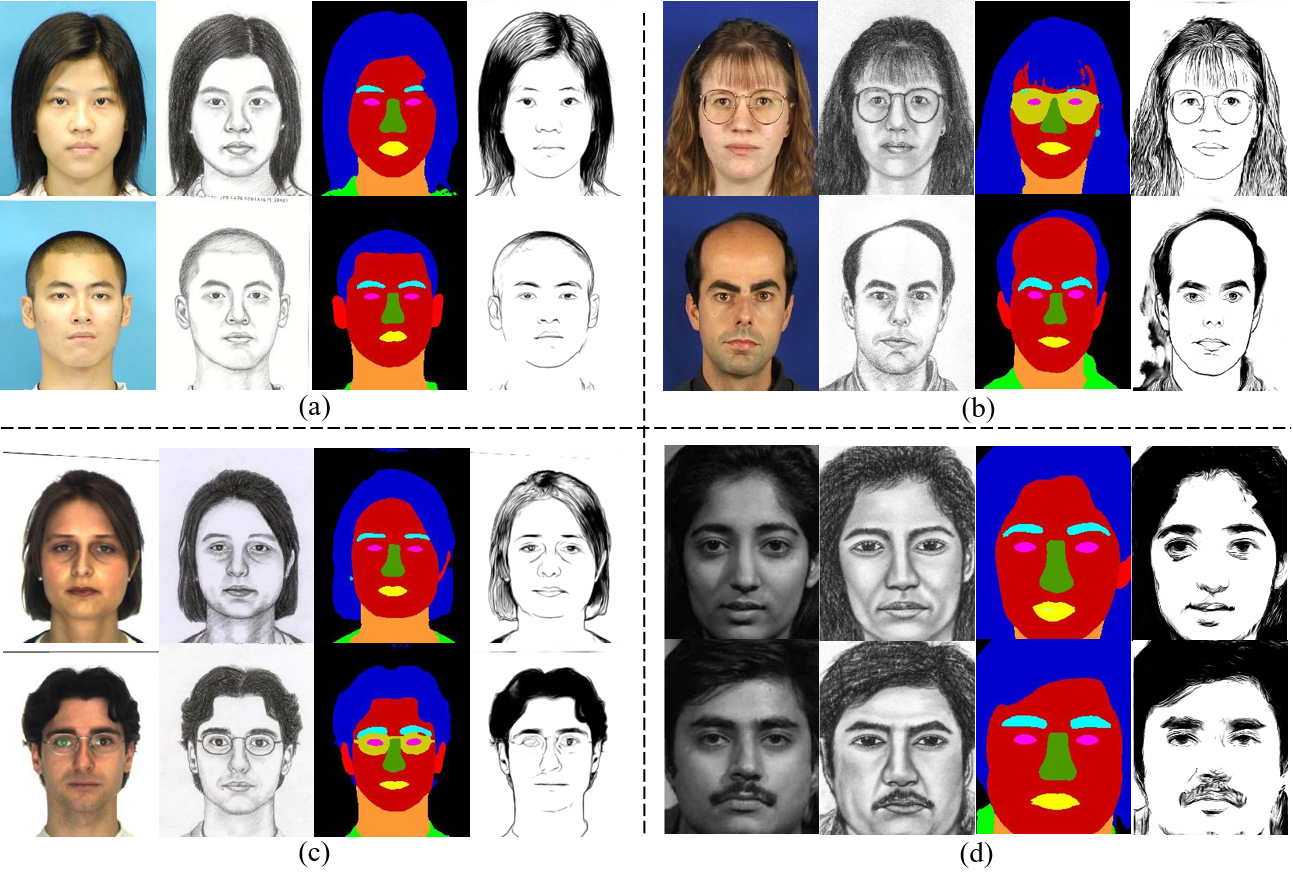}
\end{center}
   \caption{Illustrations of samples with facial prior knowledge in different databases:  (a) CUHK database, (b) XM2VTS database, (c) AR database, (d) CUFSF database. For each sample, from left to right are  face photo, sketch, face parsing layout, and saliency detection result.
}
\label{fig:prior}
\end{figure}

Exemplar-based approaches are the mainstream to solve face sketch synthesis tasks in early studies \cite{tang2003face,zhang2019deep,zhu2017deep}. They are mainly dedicated to finding the correlation between exemplar photo patches and test photo patches in the photo-sketch paired dataset \cite{zhu2017deep}. %Exemplar patches are selected from the training photo set and sketch set which should be paired.
And the final output sketch is directly reconstructed by blending the exemplar patches corresponding to the test photo patches. Although these approaches achieve favorable performance in the photo-sketch paired datasets, the generated sketches have obvious drawbacks. The synthesized sketches are too smooth to hold the fidelity and identifiability of the corresponding face photo. Moreover, exemplar-based approaches need massive computational cost and inference time in the patch-wise matching process. Exemplar-based approaches are highly dependent on exemplars and have defective effects on the sketches generated by low-quality photos. Linear-regression based approaches assume that there is a linear mapping between the face photos and sketches. Then the sketches are generated by modeling this mapping from photos \cite{liu2005nonlinear,zhou2012markov}. These approaches are mostly inspired by Locally Linear Embedding (LLE) \cite{roweis2000nonlinear} which promotes the early sketch generation task due to the lower computation complexity. However, the mapping function might not be accurately formulated, resulting in the poor quality of the generated sketches \cite{liu2005nonlinear}.
Recently, the image-to-image translation task has made great progress with the development of deep learning, especially with Generative Adversarial Network (GAN) \cite{goodfellow2014generative}.
A large number of researchers are motivated to employ GAN on sketch generation tasks \cite{zhang2015end,zhang2018face,zhu2020knowledge,yu2020toward}.
The results of the face sketch synthesis also achieve a new benchmark. However, these methods are often short of the realistic detailed depiction of the synthesized sketches.
%Though these methods have achieved decent performance in facial sketch synthesis, the generated details are not realistic and the methods are not robust enough.

Motivated by the above researches, we propose a novel Semantic-Driven Generative Adversarial Network (SDGAN) for face sketch synthesis. Firstly, we utilize the pre-trained facial saliency detection network U2-Net \cite{qin2020u2} on the input photos to obtain the prior information of overall facial texture structure.  Besides, we observe that the previous GAN networks are susceptible to illumination variations and complex backgrounds.
%of the face photos when generating the sketches.
Inspired by the great development of face parsing, we exploit the face parsing layouts as the semantic-level spatial prior to enforce globally structural style injection in the generator of SDGAN. Meanwhile, we divide face photo and sketch into local semantic classes, and propose a novel Adaptive Re-weighting Loss (ARLoss) which engages to balance the contributions of different semantic classes. Consequently, our method achieves state-of-the-art performance in a variety of metrics on CUFS and CUFSF datasets.

\begin{figure*}[t]
\begin{center}
\includegraphics[width=0.83\linewidth]{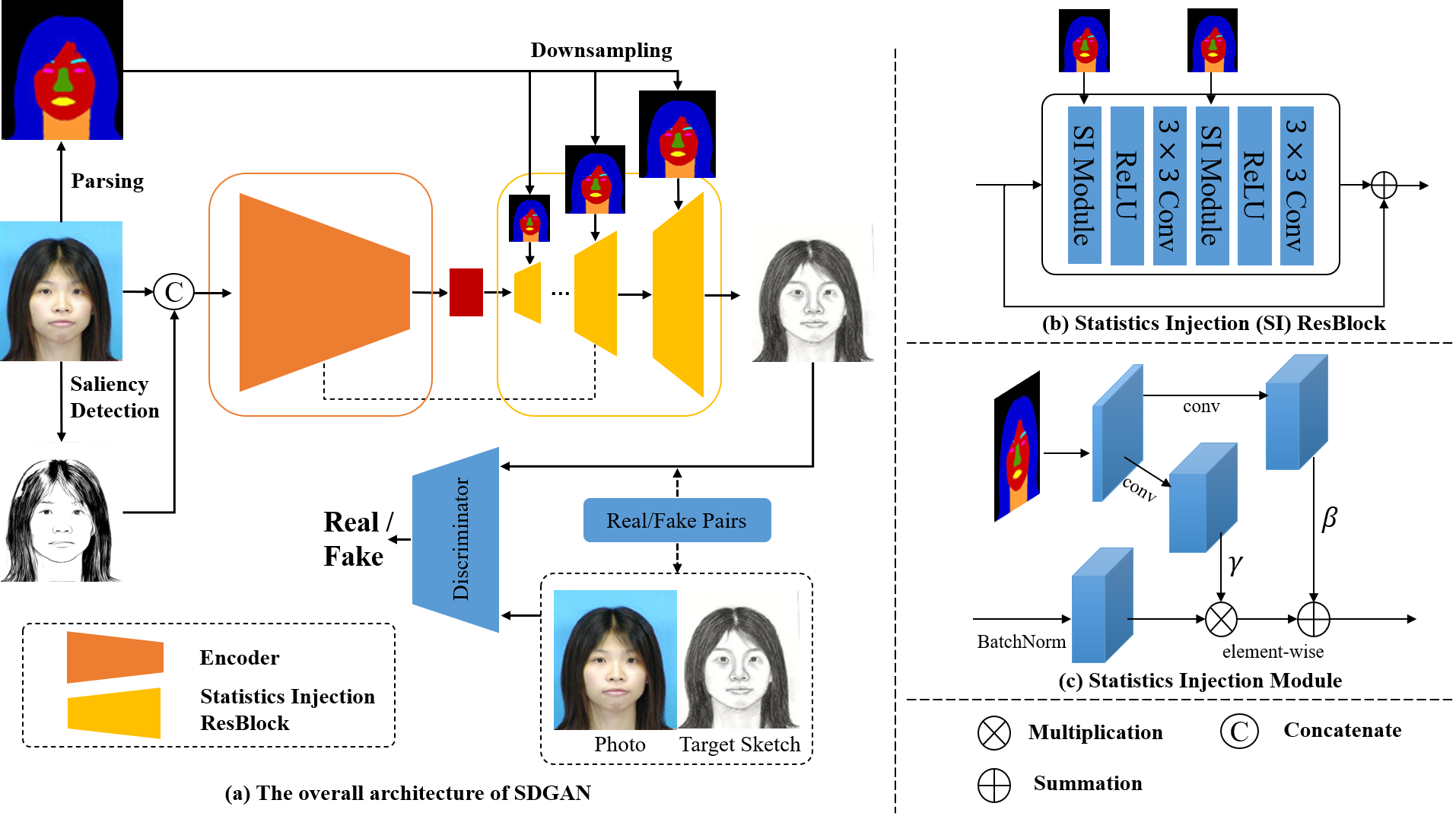}
\end{center}
   \caption{The pipeline of SDGAN. (a) The over architecture of SDGAN, which exploits Pix2Pix as backbone. The decoder contains several Statistics Injection (SI) ResBlocks with upsampling layers. (b) Details of Statistics Injection (SI) ResBlocks. (c) Details of Statistics Injection Module. In each Statistics Injection Module,  the semantic map is convoluted to produce pixel-level normalization parameters $\gamma $ and $\beta $.}
\label{fig:architecture}
\end{figure*}

The main contributions can be summarized as:

$\bullet$  We propose a novel Semantic-Driven Generative Adversarial Network for accurate and realistic face sketch synthesis.

$\bullet$  We leverage the facial saliency detection and face parsing layouts as global structural-level prior information in the generator of SDGAN to facilitate sketch synthesis.

$\bullet$  We propose a novel class-level knowledge Adaptive Re-weighting Loss (ARLoss) to balance the importance of the different semantic parts.

$\bullet$  We conduct extensive comparative experiments on CUFS and CUFSF datasets and obtain state-of-the-art performance.

\section{Related Work}

\subsection{Face Photo-Sketch Synthesis}

Face sketch synthesis has developed rapidly in the last few decades. Tang and Wang \cite{tang2003face} proposed a linear transformation method to construct face sketches by taking advantage of the Principal Component Analysis (PCA). This technique is the pioneering work of exemplar-based approaches. Linear-regression based approaches benefit from their low-cost computation and appear after exemplar-based approaches \cite{zhang2019deep,zhu2017deep}. Zhang \emph{et al.} \cite{zhang2011face} used Support Vector Regression (SVR) to obtain high-frequency information for sketch refinement. Model-based approaches are the mainstream routines of facial sketch synthesis in recent years which have gradually emerged with the boost of deep neural networks. Zhang \emph{et al.} \cite{zhang2015end} utilized a fully convolutional network to generate sketch. Ji \emph{et al.} \cite{zhang2018face} used multi-domain adversarial methods to construct a mapping from photo-domain to sketch-domain. Zhu \emph{et al.} \cite{zhu2020knowledge} borrowed knowledge from transfer learning and proposed a lightweight network supervised by a high-performance larger network. Recently, Yu \emph{et al.} \cite{yu2020toward} decomposed the face parsing layouts into multiple compositions and encoded them into cGAN for face sketch synthesis which achieved state-of-the-art performance.

\subsection{Paired Image-to-Image Translation}

The image-to-image translation is often formulated as pixel-wise image generation tasks like face sketch synthesis which are applied with paired images. Isala \emph{et al.} \cite{isola2017image} proposed a conditional GAN architecture to solve the image-to-image translation task with paired input and output named Pix2Pix. Due to the eminent performance of the Pix2Pix on the paired dataset, researchers have made numerous improvements based on Pix2Pix and applied them to a wide range of other fields \cite{qu2019enhanced,wang2018enhancing}. By combining Pix2Pix and residual blocks, Wang \emph{et al.} \cite{wang2018high} proposed a novel network architecture to generate high-resolution images named pix2pixHD. Moreover, Park \emph{et al.} \cite{park2019semantic} introduced the semantic layouts as spatial supervision injected in the pix2pixHD for synthesizing photorealistic images. Motivated by previous researches, we exploit cGAN like Pix2Pix as our backbone network.

%\subsection{Facial Prior Knowledge}

%Face parsing is a subtask of image semantic segmentation which orients toward classifying each pixel in the face image. Long \emph{et al.}\cite{long2015fully} drew on fully convolutional networks (FCN) to achieve significant breakthroughs in semantic segmentation tasks. Due to the excellent capability of FCN in the face parsing task, some researchers ponder on using the facial parsing layouts as prior knowledge in sundry facial generation tasks. Zhu \emph{et al.}\cite{zhu2020sean} made use of the decomposed facial parsing layouts to push the frontier of interactive image editing. Buhler \emph{et al.}\cite{buhler2020deepsee} leveraged the face parsing maps for extreme super-resolution task which achieved $32\times $ magnification. Besides, Yang \emph{et al.}\cite{yang2020hifacegan} formulated the face restoration as a semantic-driven task which introduced face parsing layouts to replenish facial details. In addition to face parsing, saliency detection is also a mature task in computer vision, which is dedicated to locate the important parts of images. Qin \emph{et al.}\cite{qin2020u2} designed a powerful two-level nested U-structure network named U2-Net for saliency object detection. We encode facial saliency information produced by U2-Net as prior information into the SDGAN. Moreover, the applications of face parsing layouts and facial saliency detection are shown in Fig. \ref{fig:prior}.

\subsection{Image Style Transfer}
Face sketch synthesis could be regarded as a branch of image style transfer. Gatys \emph{et al.} \cite{gatys2016image} successfully applied pre-trained CNNs to the Image style transfer task. Furthermore, Ulyanov \emph{et al.} \cite{ulyanov2016texture,ulyanov2017improved} optimized the style transfer process by manipulating the Batch Normalization (BN) layers and Instance Normalization (IN) layers. Dumoulin \emph{et al.} \cite{dumoulin2016learned} utilized a group of parameters to realize the transfer of various image styles. Consecutively, Huang \emph{et al.} \cite{huang2017arbitrary} proposed the adaptive instance normalization (AdaIN) layers which could perform arbitrary style transfer without training repeatedly. Recently, Park \emph{et al.} \cite{park2019semantic} put forward the spatially-adaptive normalization (SPADE) layers that inject the image style from the semantic layouts to obtain photorealistic images.
%------------------------------------------------------------------------

\section{Method}

%\subsection{Preliminaries}

Our semantic-driven network aims to construct a mapping from photo to sketch by utilizing semantic layouts and saliency detection. Previous researches synthesize sketches directly without taking advantage of semantic information. However, the translation from face to sketch is paired, and the maintenance of semantic information is extremely significant. Our network adopts semantic layouts to guide the generation of the sketches, especially the detailed regions. Moreover, we propose a novel Adaptive Re-weighting Loss (ARLoss) which dedicates to balance the contributions of different semantic classes. Besides, we conduct facial saliency detection on the input face photos to provide overall facial texture structure.

Given paired training photo-sketch samples $\begin{Bmatrix}
\left ( x_{i},y_{i} \right )|x_{i}\in X, y_{i}\in Y
\end{Bmatrix}_{i=1}^{N}$, where $x_{i}$ represents photo and $y_{i}$ represents sketch. The purpose of face sketch synthesis is to construct a mapping from source photo domain $X$ to target sketch domain $Y$. As illustrated in Fig. \ref{fig:prior}, we find that there are illumination variations and complex backgrounds in the source domain resulting in severe impacts on the identity and fidelity of the generated sketch. To handle these challenging issues, we first utilize face saliency detection results as prior information to provide the overall facial texture structure. We concatenate the texture structure $M$ and the photo $X$ as the input to the generator.
%Thanks to the development of the face parsing networks~\cite{lee2020maskgan,yu2018bisenet}, they show excellent performance under illumination variations and complex backgrounds. 
Besides, we also employ the pre-trained face parsing network to acquire semantic layouts $S$. Then the face semantic information $S$ is injected into our network to produce the final synthesized result. Therefore, the overall mapping can be formulated as $\begin{Bmatrix}X,M,S\end{Bmatrix}\rightarrow Y$.

%Simultaneously, we observe that the face parsing networks MaskGAN \cite{lee2020maskgan} or BiSeNet \cite{yu2018bisenet} show excellent performance under illumination variations and complex backgrounds. Therefore, we employ pre-trained MaskGAN and BiSeNet to acquire semantic layouts $S$. Then the detailed knowledge of the face structure is injected into our network. Therefore, the mapping can be formulated as $\begin{Bmatrix}X,M,S\end{Bmatrix}\rightarrow Y$.

\subsection{Network Architecture}

Fig.~\ref{fig:architecture} illustrates the overall architecture of our network.  
%We aim to construct the mapping $\begin{Bmatrix}X,M,S\end{Bmatrix}\rightarrow Y$. 
Specifically, we concatenate the paired $M$ and $X$ as inputs which are forwarded to the network to supply the overall facial texture structure.
Pix2Pix \cite{isola2017image} is exploited as the backbone which contains 7 convolutional and downsampling layers in the encoder part. 
In order to further strengthen the conditional semantic information in the forwarding process, 
we also design 7 Statistics Injection (SI) ResBlocks in the decoder, which is motivated by~\cite{park2019semantic}.
%As for the decoder, it also contains 7 Statistics Injection (SI) ResBlocks and upsampling layers.
As shown in Fig. \ref{fig:architecture} (b), the SI ResBlock consists of two convolutional layers, two ReLU layers, and two Statistics Injection (SI) modules.
%Motivated by SPADE \cite{park2019semantic}, we design a Statistics Injection (SI) module for semantic injection.

As shown in Fig. \ref{fig:architecture} (c), each SI module takes two inputs: the forward activation features after Batch Normalization layer and semantic masks $S$, which are obtained by pre-trained MaskGAN \cite{lee2020maskgan} or BiSeNet \cite{yu2018bisenet}.
In order to prevent semantic ambiguity, we merge all facial parts into 12 classes which are closely related to sketches: two eyes, two eyebrows, two ears, glasses, upper and lower lips, inner mouth, hair, nose, skin, neck, cloth, and background.
%%In the decoder part, we employ SPADE ResBlocks\cite{park2019semantic}%%
\begin{figure}
\begin{center}
\includegraphics[width=0.95\linewidth]{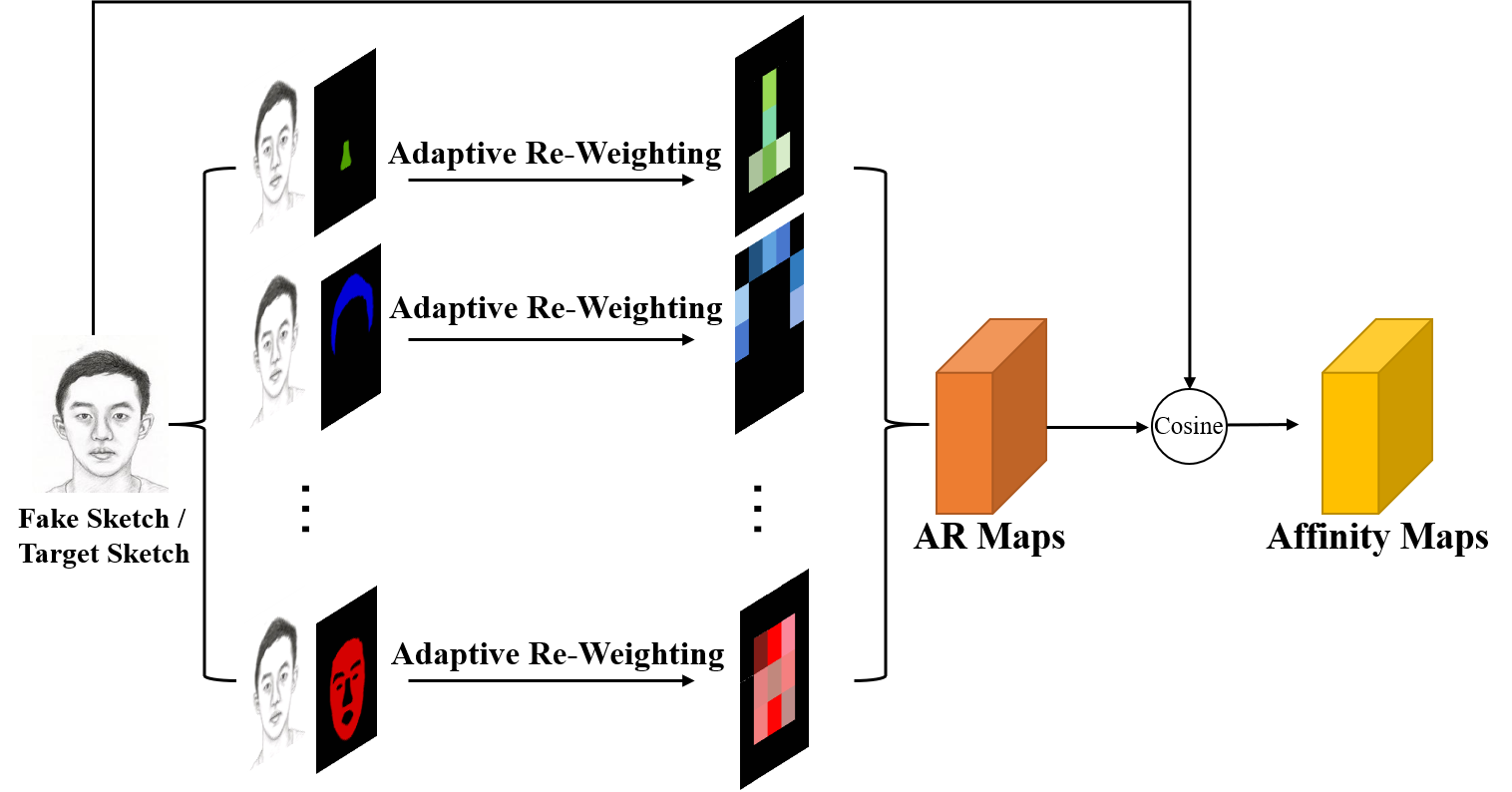}
\end{center}
   \caption{Illustration of Adaptive Re-Weighting. The semantic masks are produced by pre-trained face parsing network from the input photo. The AR maps consist of $\mu$ and $\nu$ with spatial dimension = 24 (12+12) in practice. }
\label{fig:AR}
\end{figure}
Therefore, we have $S= \begin{Bmatrix}s^{(1)},\cdots,s^{(c)} \end{Bmatrix} \in \mathbb{R}^{h\times w\times c}$, where $c\in \begin{bmatrix}1,2,\cdots,12\end{bmatrix}$, $s^{(c)}\in \begin{bmatrix}0,1\end{bmatrix}$, $h$ and $w$ denote the height and width of the feature maps.
%``1'' represents the spatial content of a specific class, and ``0'' represents other classes. 
In order to inject the semantic information into the SI module, we perform the convolutional operation on $S$ to produce the modulation parameters $\gamma $ and $\beta $ to normalize the final output. The produced $\gamma$ and $\beta$ encode sufficient spatial layout information which multiplied and added to the normalized activation through an element-wise way as shown in Fig. \ref{fig:architecture} (c). In fact, the modulation parameters  $\gamma $ and $\beta $ could provide a kind of spatial supervision from $X$ to $Y$ through $S$ which are robust to illumination variations and complex backgrounds.
Finally, the network structure of the discriminator keeps the same settings as Pix2Pix.
%Semantic masks $S$ could be treated as the intermediate domain from source domain $X $ to target domain $Y$.
%Consequently, the modulation parameters  $\gamma $ and $\beta $ could provide a kind of spatial supervision from $X$ to $Y$ through $S$. Additionally, the network structure of the discriminator keeps the same settings as Pix2Pix.

\begin{figure*}[t]
\begin{center}
\includegraphics[width=0.87\linewidth]{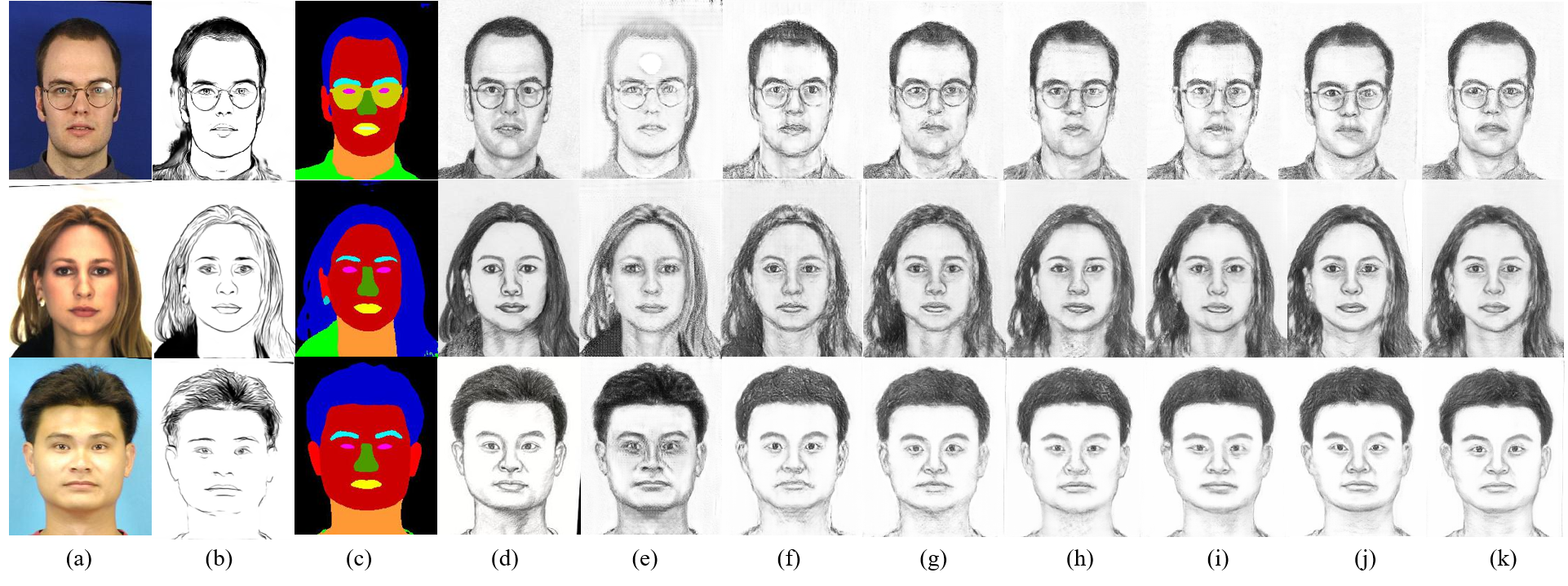}
\end{center}
   \caption{Ablation studies of synthesized sketches on the CUFS dataset. From the top to bottom, the examples are selected from XM2VTS database, AR database, and CUHK database. (a) Photo, (b) Saliency detection map $M$, (c) Semantic parsing mask $S$, (d) Target sketch, (e) CycleGAN \cite{zhu2017unpaired}, (f) Pix2Pix \cite{isola2017image}, (g) SDGAN w/ $M$ , (h) SDGAN w/ $M$ + $S$, (i) SDGAN w/ $M$ + $S$ + ARLoss, (j) SDGAN w/ $M$ + $S$ + ARLoss + Perceptual Loss, (k) SDGAN w/ $M$ + $S$ + ARLoss + Perceptual Loss + BCE Loss. }
\label{fig:CUFS}
\end{figure*}

\subsection{Adaptive Re-weighting}
In previous studies, researchers always impose overall supervision and constraints on the entire generated sketches, which leads to defective performances in facial details. In this paper, we propose an Adaptive Re-weighting algorithm which could trade-off the contributions of different semantic classes, especially the detailed parts. As illustrated in Fig.~\ref{fig:AR}, the synthesized sketch is represented as $F\in R^{h_{f}\times w_{f}\times c_{f}}$, where $h_{f}$, $w_{f}$ and $c_{f}$ denote the height, width and channel of the sketch, respectively. Then we enforce the element-wise multiplication between $F$ and semantic masks $S$ to extract the Interest-Region (IR) for each semantic class. The Adaptive Re-weighting could be formulated as:
\begin{align}
\mu\begin{pmatrix}
c
\end{pmatrix} = \frac{1}{\begin{vmatrix}
S\begin{pmatrix}
:,:,c
\end{pmatrix}
\end{vmatrix}}\sum_{i=1}^{h_{f}}\sum_{j=1}^{w_{f}}S\begin{pmatrix}
i,j,c
\end{pmatrix}F\begin{pmatrix}
i,j
\end{pmatrix}.
\end{align}
where  ${\begin{vmatrix}S\begin{pmatrix}:,:,c\end{pmatrix}\end{vmatrix}}$ represents the summation of pixel numbers in each IR with the same semantic class $c$. Obviously, this strategy normalizes each IR with ${\begin{vmatrix}S\begin{pmatrix}:,:,c\end{pmatrix}\end{vmatrix}}$ which could balance the contributions of different semantic classes. Besides, $\mu(c)$ also could be considered as the mean value of all pixels in the $c-th$ semantic category. Furthermore, we introduce the modulation variance $\nu(c)$ to faithfully react the semantic variation of the intra-class feature distribution. Formally, the computation is listed as follows:
\begin{align}
\nu\begin{pmatrix}
c
\end{pmatrix} = \frac{1}{\begin{vmatrix}
S\begin{pmatrix}
:,:,c
\end{pmatrix}
\end{vmatrix}}\sum_{i=1}^{h_{f}}\sum_{j=1}^{w_{f}}\left \{ S\begin{pmatrix}
i,j,c
\end{pmatrix}F\begin{pmatrix}
i,j
\end{pmatrix}-\mu \begin{pmatrix}
c
\end{pmatrix} \right \}^{2}.
\end{align}

Note that both $\mu$ and $\nu$ are tensors. We named $\mu$ and $\nu$ as Adaptive Re-weighting (AR) maps. Finally, we construct the affinity maps between the synthesized sketch $F$ and $\mu$ ( or $\nu$), which could be calculated by Cosine similarity.
 \begin{align}
 \mathbb{C}_{1}=&\frac{F\cdot \mu }{\left \| F \right \|_{2}\cdot \left \| \mu  \right \|_{2}}
\notag
 \\\mathbb{C}_{2}=&\frac{F\cdot \nu }{\left \| F \right \|_{2}\cdot \left \| \nu  \right \|_{2}}.
\end{align}
Therefore, the synthesized sketch and the target sketch re-weight each semantic class of IR by constructing AR maps. 

%Furthermore, we design an AR Loss through this strategy, which aims at performing AR on the synthesized sketch and target sketch and then constrain the $L_{2}$ distance, as described in Section 3.4 (C).

\subsection{Objective Function}
The overall objective of our model includes five loss functions: $\mathcal{L}_{GAN}$, $\mathcal{L}_{content}$, $\mathcal{L}_{AR}$, $\mathcal{L}_{perceptual}$ and $\mathcal{L}_{BCE}$.

\textbf{\emph{A) Adversarial Loss.}} The adversarial loss is leveraged to correctly distinguish the real sketches or generated  sketches. Follows the setting of \cite{isola2017image}, the adversarial loss is formulated as:
\begin{align}
\mathcal{L}_{GAN}&=\mathit{E}_{X,M,Y}\left [ \log D\left ( X,M,Y \right ) \right ]
\notag
\\&+{E}_{X,M}\left [ \log \left ( 1-D\left ( X,M,G\left ( X,M \right ) \right ) \right ) \right ].
\end{align}
where $X$, $Y$ and $M$ denote the source photos, target sketches and saliency detection maps.

\textbf{\emph{B) Content Loss.}} In addition, we utilize the normalized $L_{1}$ distance to represent content loss since it causes less blurring than $L_{2}$ distance.
\begin{align}
\mathcal{L}_{content}\left ( G \right )=\mathit{E}_{X,M,Y}\left [ \left \| Y-G\left ( X,M \right ) \right \|_{1} \right ].
\end{align}

\textbf{\emph{C) Adaptive Re-weighting Loss.}} In practice, we extract the affinity maps from target sketch and synthesized sketch, respectively. These affinity maps contain comprehensive knowledge between sketch and class-wise re-weighting maps which represent the intra-class feature distribution. Furthermore, in the process of calculating AR map, the contribution of each semantic class is adaptively re-weighted. Consequently, we reinforce the supervision on these affinity maps to constrain the generated sketch matching the feature distribution of the target domain. The Adaptive Re-weighting (AR) loss is formulated as:
\begin{align}
\mathcal{L}_{AR}\left ( \mathbb{C}^{target}, \mathbb{C}^{fake} \right )&=\\
\notag
&\sum_{r=1}^{2}\sum_{c=1}^{12}\left \| \mathbb{C}_{r}^{target}\left ( c \right ) -\mathbb{C}_{r}^{fake}\left ( c \right )\right \|_{2}^{2}.
\end{align}
where the $r \in \begin{bmatrix}1,2\end{bmatrix}$ denotes two types of affinity maps.

\textbf{\emph{D) Perceptual Loss.}}  In order to ensure that the generated sketch and the target sketch have more similar specificity, we employ pre-trained VGG-19 net \cite{simonyan2014very} as feature extractor to obtain high-level representations. We compare the features from VGG-19 after pool1 and pool2 layers. The perceptual loss engages to make training procedure more stable.
\begin{align}
\mathcal{L}_{perceptual}=\sum_{l=1}^{2}\left \| \omega^{l} \left ( Y \right ) -\omega^{l} \left ( G\left ( X,M \right ) \right )\right \|_{2}^{2}.
\end{align}
where $\omega ^{l}\left ( \cdot  \right )$ represents the output features of VGG-19 net and $l$ denotes the selected pool1 and pool2 layers.

\textbf{\emph{E) Binary Cross-Entropy Parsing Loss.}}Finally, we introduce the Binary Cross-Entropy (BCE) loss to further refine the synthesized sketch in semantic level. We contrast the semantic mask of synthesized sketch and target sketch produced by pre-trained parsing network \cite{lee2020maskgan,yu2018bisenet}.
\begin{align}
\mathcal{L}_{BCE}=\left ( \mathbb{P}\left ( Y \right ), \mathbb{P}\left ( G\left ( X,M \right ) \right )\right ).
\end{align}
where $\mathbb{P}$ denotes the inference process of parsing network.

\textbf{\emph{F) Full Objective.}} Eventually, we combine all loss functions to achieve overall supervision:
\begin{align}
\mathcal{L}_{total} = \mathcal{L} _{GAN}&+\alpha \mathcal{L}_{content}+\lambda \mathcal{L} _{AR}
\notag
\\&+\delta \mathcal{L} _{perceptual}+\eta \mathcal{L} _{BCE}.
\end{align}
where the $\alpha $, $\lambda$, $\delta$ and $\eta$ are weighting factors. Furthermore, the generator $G$ and the discriminator $D$ could be optimized by the following formulation:
\begin{align}
\min_{G}\max_{D}\mathcal{L}_{total}
\end{align}

\begin{table*}
\begin{center}
\resizebox{\textwidth}{!}{
\renewcommand{\arraystretch}{1.15} % default is 1.0
\begin{tabular}{|p{2.6cm}|p{1.2cm}<{\centering} p{1.2cm}<{\centering}| p{1.2cm}<{\centering} p{1.2cm}<{\centering}| p{1.2cm}<{\centering} p{1.2cm}<{\centering}| p{1.2cm}<{\centering} p{1.2cm}<{\centering}| p{1.2cm}<{\centering} p{1.25cm}<{\centering}| p{1.2cm}<{\centering} p{1.2cm}<{\centering}| p{1.2cm}<{\centering} p{1.2cm}<{\centering} |}
\hline
\textbf{Methods / Years}& \multicolumn{2}{c}{\textbf{FSIM}$\blacktriangle$ }& \multicolumn{2}{c}{\textbf{SSIM}$\blacktriangle$} & \multicolumn{2}{c}{\textbf{FID}$\blacktriangledown$ }& \multicolumn{2}{c}{\textbf{LPIPS (SqueezeNet)}$\blacktriangledown$} & \multicolumn{2}{c}{\textbf{LPIPS (AlexNet)}$\blacktriangledown$} &
\multicolumn{2}{c|}{\textbf{LPIPS (VGG-16)}$\blacktriangledown$ }\\
\cline{2-13}
\textbf{Datasets} & \textbf{CUFS} & \textbf{CUFSF} & \textbf{CUFS} &\textbf{CUFSF} & \textbf{CUFS} & \textbf{CUFSF} & \textbf{CUFS} & \textbf{CUFSF} & \textbf{CUFS} & \textbf{CUFSF} & \textbf{CUFS} & \textbf{CUFSF}\\
\hline
CycleGAN \cite{zhu2017unpaired}  & \multirow{2}{*}{0.6829} & \multirow{2}{*}{0.7011} & \multirow{2}{*}{0.4638} & \multirow{2}{*}{0.3753} & \multirow{2}{*}{58.394} & \multirow{2}{*}{31.262} & \multirow{2}{*}{0.1863} & \multirow{2}{*}{0.1617} & \multirow{2}{*}{0.2776} & \multirow{2}{*}{0.2234 }& \multirow{2}{*}{0.3815} & \multirow{2}{*}{0.3787} \\
(2017) &~ &~ &~ &~ &~ &~ &~ &~ &~ &~&~&~ \\
\hline
Pix2Pix \cite{isola2017image} (2017) & 0.7356 & 0.7284 & 0.5172 & 0.4204 & 44.272 & 30.984 & 0.1156 & 0.1422 & 0.1654 & 0.1932 & 0.3059 & 0.3551\\
\hline
MDAL \cite{zhang2018face} & \multirow{2}{*}{0.7275} & \multirow{2}{*}{0.7076 }& \multirow{2}{*}{0.5280} &  \multirow{2}{*}{0.3818} &\multirow{2}{*}{/} & \multirow{2}{*}{/}& \multirow{2}{*}{/} & \multirow{2}{*}{/} &\multirow{2}{*}{/} & \multirow{2}{*}{/} & \multirow{2}{*}{/}& \multirow{2}{*}{/}\\
(2018)  &~ &~ &~ &~ &~ &~ &~ &~ &~ &~&~&~ \\
\hline
KT \cite{zhu2019face} (2019) & 0.7373 & 0.7311 & / & / & / & / & 0.1688 & 0.1740 & 0.2297 & 0.2522 & 0.3483 & 0.3743\\
\hline
Col-cGAN \cite{zhu2019deep}& \multirow{2}{*}{/} & \multirow{2}{*}{/} & \multirow{2}{*}{0.5244} & \multirow{2}{*}{0.4224}  & \multirow{2}{*}{/} &\multirow{2}{*}{/} & \multirow{2}{*}{/} &\multirow{2}{*}{/} & \multirow{2}{*}{/} &\multirow{2}{*}{/} &\multirow{2}{*}{/} &\multirow{2}{*}{/}\\
(2019)&~ &~ &~ &~ &~ &~ &~ &~ &~ &~&~&~\\
\hline
KD+ \cite{zhu2020knowledge} (2020) & 0.7350 & 0.7171 & / & / & / & / & 0.1471 & 0.1619 & 0.1971 & 0.2368 & 0.3052 & 0.3550\\
\hline
MSG-SARL \cite{duan2020multi}  & \multirow{2}{*}{\textbf{0.7594}} & \multirow{2}{*}{0.7316} & \multirow{2}{*}{0.5288} & \multirow{2}{*}{0.4230} & \multirow{2}{*}{46.39 }& \multirow{2}{*}{38.25 }& \multirow{2}{*}{/} &\multirow{2}{*}{/} & \multirow{2}{*}{/} &\multirow{2}{*}{/} & \multirow{2}{*}{/} &\multirow{2}{*}{/}\\
(2020)&~ &~ &~ &~ &~ &~ &~ &~ &~ &~&~&~\\

\hline
SCAGAN \cite{yu2020toward} & \multirow{2}{*}{0.716} & \multirow{2}{*}{0.729} & \multirow{2}{*}{/} & \multirow{2}{*}{/} & \multirow{2}{*}{34.2} & \multirow{2}{*}{\textbf{18.2}} &\multirow{2}{*}{ /} &\multirow{2}{*}{ /} &\multirow{2}{*}{/} &\multirow{2}{*}{/} & \multirow{2}{*}{/ }& \multirow{2}{*}{/}\\
(2020) &~ &~ &~ &~ &~ &~ &~ &~ &~ &~&~&~\\
\hline
\textbf {SDGAN(ours)} & 0.7446& \textbf{0.7328} & \textbf{0.5360} & \textbf{0.4339 }& \textbf{33.408} & 30.594 & \textbf{0.1017}& \textbf{0.1370} & \textbf{0.1444}& \textbf{0.1906} & \textbf{0.2767}& \textbf{0.3358}\\
\hline
\end{tabular}}
\end{center}
\caption{\label{tab:CUFS}Comparison of different models in the CUFS dataset and the CUFSF dataset. $\blacktriangle$ indicates the higher is better, $\blacktriangledown$ indicates the lower is better. Our method reaches the \textbf{\emph{optimal}} and \textbf{\emph{sub-optimal}} results in the CUFS dataset and the CUFSF dataset.  }
\end{table*}

%------------------------------------------------------------------------

\begin{figure*}[t]
\begin{center}
\includegraphics[width=0.87\linewidth]{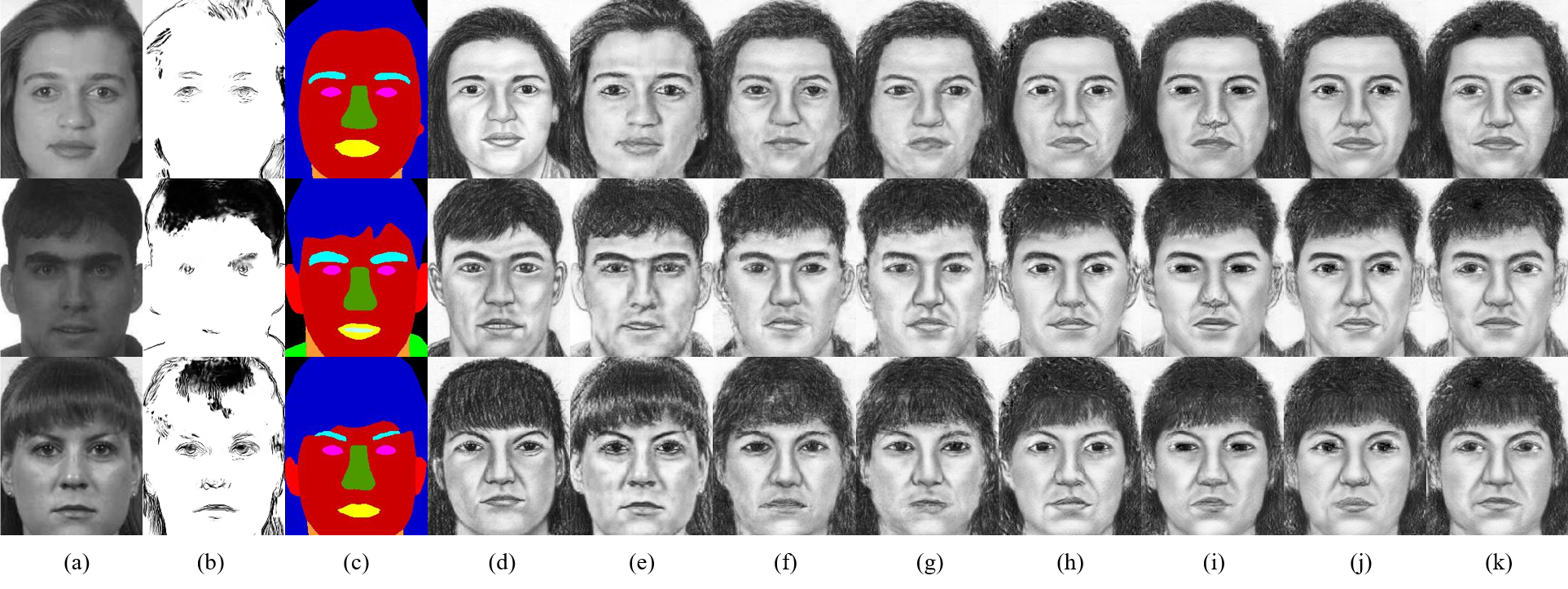}
\end{center}
   \caption{Ablation studies of synthesized sketches on the CUFSF dataset. (a) Photo, (b) Saliency detection map $M$, (c) Semantic parsing mask $S$, (d) Target sketch, (e) CycleGAN \cite{zhu2017unpaired}, (f) Pix2Pix \cite{isola2017image}, (g) SDGAN w/ $M$ , (h) SDGAN w/ $S$, (i) SDGAN w/ $S$ + ARLoss, (j) SDGAN w/ $S$ + ARLoss + Perceptual Loss, (k) SDGAN w/ $S$ + ARLoss + Perceptual Loss + BCE Loss. }
\label{fig:CUFSF}
\end{figure*}

\begin{table*} [htbp]
\begin{center}
\resizebox{\textwidth}{!}{
\renewcommand{\arraystretch}{1.15} % default is 1.0
\begin{tabular}{|p{3cm}|p{1.2cm}<{\centering} p{1.2cm}<{\centering}| p{1.2cm}<{\centering} p{1.2cm}<{\centering}| p{1.2cm}<{\centering} p{1.2cm}<{\centering}| p{1.2cm}<{\centering} p{1.2cm}<{\centering}| p{1.2cm}<{\centering} p{1.25cm}<{\centering}| p{1.2cm}<{\centering} p{1.2cm}<{\centering}| p{1.2cm}<{\centering} p{1.2cm}<{\centering} |}
\hline
\textbf{Methods /}& \multicolumn{2}{c}{\textbf{FSIM}$\blacktriangle$ }& \multicolumn{2}{c}{\textbf{SSIM}$\blacktriangle$} & \multicolumn{2}{c}{\textbf{FID}$\blacktriangledown$ }& \multicolumn{2}{c}{\textbf{LPIPS (SqueezeNet)}$\blacktriangledown$} & \multicolumn{2}{c}{\textbf{LPIPS (AlexNet)}$\blacktriangledown$} &
\multicolumn{2}{c|}{\textbf{LPIPS (VGG-16)}$\blacktriangledown$ }\\
\cline{2-13}
\textbf{Datasets} & \textbf{CUFS} & \textbf{CUFSF} & \textbf{CUFS} &\textbf{CUFSF} & \textbf{CUFS} & \textbf{CUFSF} & \textbf{CUFS} & \textbf{CUFSF} & \textbf{CUFS} & \textbf{CUFSF} & \textbf{CUFS} & \textbf{CUFSF}\\
\hline
Backbone & 0.7356 & 0.7284 & 0.5172 & 0.4204 & 44.272 & 30.984 & 0.1156 & 0.1422 & 0.1654 & 0.1932 & 0.3059 & 0.3551\\
\hline
SDGAN w/ $M$ & 0.7376 & 0.7275 & 0.5145 & 0.4193 & 42.367 & 29.765 & 0.1141 & 0.1429 & 0.1581 & 0.1939 & 0.3080 & 0.3562\\
\hline
SDGAN w/ $M$ + $S$ & 0.7411 & 0.7290 & 0.5286 & 0.4297 & 41.048 & 30.970 & 0.1108 & 0.1443& 0.1570 & 0.2037 & 0.2965 & 0.3454\\
\hline
SDGAN w/ $M$ + $S$ &\multirow{2}{*}{0.7421} & \multirow{2}{*}{0.7309} & \multirow{2}{*}{0.5306} & \multirow{2}{*}{0.4339} & \multirow{2}{*}{34.275} & \multirow{2}{*}{28.336} & \multirow{2}{*}{0.1078} & \multirow{2}{*}{0.1448 }& \multirow{2}{*}{0.1529} & \multirow{2}{*}{0.2054} & \multirow{2}{*}{0.2907} & \multirow{2}{*}{0.3499} \\
+ ARLoss &~ &~ &~ &~ &~ &~ &~ &~ &~ &~&~&~ \\
\hline
SDGAN w/ $M$ + $S$ + &~ &~ &~ &~ &~ &~ &~ &~ &~ &~&~&~\\ARLoss + Perceptual & 0.7433 & 0.7317 & 0.5330 & \textbf{0.4355} & \textbf{32.040} & \textbf{28.212 }& 0.1019 & 0.1407 & 0.1467 & 0.1944 & 0.2784 & 0.3424 \\
Loss &~ &~ &~ &~ &~ &~ &~ &~ &~ &~&~&~ \\
\hline
SDGAN w/ $M$ + $S$ + &~ &~ &~ &~ &~ &~ &~ &~ &~ &~&~&~\\
ARLoss + Perceptual & \textbf{0.7446} & \textbf{0.7328} & \textbf{0.5360} & 0.4339 & 33.408 & 30.594 & \textbf{0.1017}& \textbf{0.1370} & \textbf{0.1444}& \textbf{0.1906} & \textbf{0.2767}& \textbf{0.3358 }\\
Loss + BCE Loss &~ &~ &~ &~ &~ &~ &~ &~ &~ &~&~&~\\
\hline
\end{tabular}}
\end{center}
\caption{\label{tab:ablation}Ablation studies of SDGAN in the CUFS dataset and the CUFSF dataset. $\blacktriangle$ indicates the higher is better, $\blacktriangledown$ indicates the lower is better. Note that we discard $M$ after injecting the parsing semantic layouts $S$ in the experiments of the CUFSF dataset.}
\end{table*}

\section{Experiments}
\subsection{Implements Details}
Our network is trained from scratch. Both the generator and discriminator are implemented on the platform Pytorch \cite{paszke2017automatic} with a single NVIDIA GeForce Titan X GPU. We leverage the Adam optimizer with $\beta _{1}=0.5$ and $\beta _{2}=0.999$. The total training epochs are 200, then the initial learning rate is set to 0.0002 for the first 100 epochs and decay linearly in the last 100 epochs. Additionally, we utilize the Instance Normalization \cite{ulyanov2017improved}, and set the batchsize = 1. Meanwhile, the weighting factors are set as $\alpha=100$, $\lambda=100$, $\delta=1$ and $\eta=10$.

\subsection{Datasets and Evaluation Criteria}
In this article, we conduct extensive experiments on the CUHK Face Sketch Dataset (CUFS) \cite{tang2003face} and the CUHK Face Sketch FERET Dataset (CUFSF) \cite{zhang2011coupled}. In CUFS dataset, there are 606 faces, of which 188 faces from the CUHK student database, 123 faces from the AR database, and 295 faces from the XM2VTS database. For each sample, there are paired face photo and sketch drawn by the artist in natural lighting conditions. The CUFSF dataset contains 1194 face photos with paired sketches. However, all the photos in the CUFSF dataset are acquired under illumination variations as illustrated in Fig. \ref{fig:CUFSF} (a). For the CUFS dataset and CUFSF dataset, the whole images are cropped to $200 \times 250$.  We adapt the reshaping and padding strategy in \cite{yu2020toward} to expand the input image size to $256 \times 256$.

The experimental performance is measured by multiple metrics. We employ the Feature Similarity Index Metric (FSIM) \cite{zhang2011fsim} and Structural Similarity Index Metric (SSIM) \cite{wang2004image} to evaluate the quality of synthesized sketches. Besides, we utilize the Fréchet Inception Distance (FID) \cite{heusel2017gans} to compute the distance of distributions between the target sketch domain and the synthesized sketch domain. Furthermore, we introduce the Learned Perceptual Image Patch Similarity (LPIPS) \cite{zhang2018unreasonable} to calculate the distance of embedding features between the generated sketches and target sketches. In this paper, LPIPS is exploited by three classification networks which are SqueezeNet \cite{iandola2016squeezenet} AlexNet \cite{krizhevsky2012imagenet}, and VGGNet \cite{simonyan2014very}.

\subsection{Results and Ablation Study}

\textbf{\emph{A) Results on CUFS Dataset.}} Table. \ref{tab:CUFS} shows the comparison results between our network and other state-of-the-art models in the CUFS dataset. We obtain the best performance on the indicators of SSIM, FID, LPIPS (SqueezeNet), LPIPS (AlexNet), and LPIPS (VGG-16). Our method increases the previous best SSIM from 0.5288 to 0.5360 and decreases the previous best FID from 34.2 to 33.4 in the CUFS dataset. As illustrated in Fig. \ref{fig:CUFS}, our network could integrally construct the details of each sketch with various backgrounds. The sketches we generated have the prolific identify features in the three sub-datasets of CUFS, especially in facial details.

\textbf{\emph{B) Results on CUFSF Dataset.}} It is more challenging to synthesize sketches from the CUFSF dataset.  As shown in Fig. \ref{fig:CUFSF} (a), there are illumination variations in the CUFSF dataset. Meanwhile, all photos are grayscale. However, our network achieves prodigious performance on the CUFSF dataset. We obtain the best sores on the FSIM, SSIM, LPIPS (SqueezeNet), LPIPS (AlexNet), and LPIPS (VGG-16) as shown in Table. \ref{tab:CUFS}. Specifically, our network increases the previous best FSIM from 0.7316 to 0.7328 and increases the previous best SSIM from 0.4230 to 0.4339 in the CUFSF dataset.

Ultimately, our results achieve optimal and sub-optimal in the CUFS dataset and CUFSF dataset.

\textbf{\emph{C) Ablation Study.}}
Furthermore, we carry out abundant ablation studies to prove the capability of our proposed Adaptive Re-weighting algorithm. We employ the saliency detection map $M$ as prior information which faithfully improves the overall facial texture structure of synthesized sketches in the CUFS dataset. Additionally, we exploit the pre-trained MaskGAN \cite{lee2020maskgan} to extract the semantic layouts. Crucially, we generate the more realistic sketches by performing spatial supervision and re-weighting the 12 semantic parts of the face. Then, we utilize Perceptual Loss and Binary Cross-Entropy Loss to further refine the facial details of the generated sketch. As shown in Fig. \ref{fig:CUFS} (j, k) and Fig \ref{fig:CUFSF} (j, k), the detailed parts (like mouth, hair, and eyes) of the sketches are more vivid. The fly in the ointment is that we observe the saliency detection map $M$ affects the performance of our network because of its low quality as depicted in Fig. \ref{fig:CUFSF} (b). Consequently, we discard $M$ as the prior input to the network in the CUFSF dataset. Furthermore, we also take the advantages of BiSeNet \cite{yu2018bisenet} to obtain more accurate semantic layouts on the different datasets.
%In addition, we find that due to the images of the CUFSF dataset are grayscale, the parsing effects of MaskGAN are seriously affected.  We employ pretrained BiSeNet\cite{yu2018bisenet} on the CUFSF dataset to obtain more accurate semantic masks in the inference process of parsing.

\section{Conclusion}
In this paper, we propose a Semantic-Driven Generative Adversarial Network (SDGAN) for face sketch synthesis by utilizing saliency detection and face parsing layouts as prior information.
Specifically, we employ a semantic-injection method and propose a novel Adaptive Re-weighting strategy which dedicates to balance the contributions of different semantic classes.
We conduct extensive experiments on the CUFS dataset and the CUFSF dataset.
Eventually, our proposed SDGAN achieves state-of-the-art performance on these two datasets. Additionally, we will conduct more experiments on the generation of faces from sketches.
A more complete version of this research will be released in the future.
%Additionally, we will apply extensive experiments which engage to synthesis the photos from sketches to verifies the capability and robustness of our network.

\subsection*{Acknowledgment}

This work was supported in part by the Natural Science Foundation of China under Grant No. 62076240, Grant No. U1836217, Grant No. 61721004; and in part by the Shandong Provincial Key Research and Development Program (Major Scientific and Technological Innovation Project) (Grant No. 2019JZZY010119)

{\small
\bibliographystyle{ieee}
\bibliography{ijcb2021_template}
}

\end{document}